\documentclass{article}
\usepackage[preprint]{spconf}
\usepackage{graphicx}
\usepackage{soul}
\usepackage{url}
\usepackage{subcaption}
\usepackage{amsmath,bm,bbm,amsthm}
\usepackage{amssymb}
\usepackage{float}
\usepackage{graphicx}
\usepackage{makecell}
\usepackage{xcolor}
\usepackage{algorithm}\usepackage[noend]{algpseudocode}
\usepackage{setspace}
\usepackage{wrapfig}
\usepackage[export]{adjustbox}
\usepackage{array}
\newcolumntype{H}{>{\setbox0=\hbox\bgroup}c<{\egroup}@{}}
\usepackage{multirow}
\usepackage{wrapfig}
\usepackage{enumitem}

\usepackage{tikz}
\usetikzlibrary{positioning}

\makeatletter
\@addtoreset{subfigure}{row}
\makeatother

\allowdisplaybreaks

\usepackage[utf8]{inputenc} 
\usepackage[T1]{fontenc}    
\usepackage{hyperref}       
\usepackage{url}            
\usepackage{booktabs}       
\usepackage{amsfonts}       
\usepackage{nicefrac}       
\usepackage{microtype}      

\newcommand{\E}{\mathbb{E}}

\newcommand{\boldI}{\boldsymbol{I}}

\newcommand{\boldP}{\boldsymbol{P}}

\newcommand{\boldV}{\boldsymbol{V}}
\newcommand{\boldW}{\boldsymbol{W}}
\newcommand{\boldX}{\boldsymbol{X}}

\newcommand{\boldb}{\boldsymbol{b}}
\newcommand{\boldd}{\boldsymbol{d}}

\newcommand{\boldv}{\boldsymbol{v}}

\newcommand{\boldx}{\boldsymbol{x}}

\newcommand{\boldz}{\boldsymbol{z}}

\newcommand{\calC}{\mathcal{C}}
\newcommand{\calD}{\mathcal{D}}

\newcommand{\calL}{\mathcal{L}}

\newcommand{\calN}{\mathcal{N}}

\newcommand{\calS}{\mathcal{S}}

\newcommand{\bLambda}{\boldsymbol{\Lambda}}

\newcommand{\btheta}{\boldsymbol{\theta}}

\title{Learning Active Subspaces for Effective and Scalable Uncertainty Quantification in Deep Neural Networks}

\name{Sanket Jantre$^{\star}$ \qquad Nathan M. Urban$^{\star}$ \qquad Xiaoning Qian$^{\star \dagger}$ \qquad Byung-Jun Yoon$^{\star \dagger}$}
\address{$^{\star}$ Computational Science Initiative, Brookhaven National Laboratory, Upton, NY \\
$^{\dagger}$ Department of Electrical \& Computer Engineering, Texas A\&M University, College Station, TX}

\copyrightnotice{\copyright This work has been submitted to the IEEE for possible publication. Copyright may be transferred without notice.}

\begin{document}
%
\maketitle
\begin{abstract}
Bayesian inference for neural networks, or Bayesian deep learning, has the potential to provide well-calibrated predictions with quantified uncertainty and robustness. However, the main hurdle for Bayesian deep learning is its computational complexity due to the high dimensionality of the parameter space. In this work, we propose a novel scheme that addresses this limitation by constructing a low-dimensional subspace of the neural network parameters--referred to as an \textit{active subspace}--by identifying the parameter directions that have the most significant influence on the output of the neural network. We demonstrate that the significantly reduced active subspace enables effective and scalable Bayesian inference via either Monte Carlo (MC) sampling methods, otherwise computationally intractable, or variational inference. Empirically, our approach provides reliable predictions with robust uncertainty estimates for various regression tasks.
\end{abstract}

%

\vspace{-0.1in}
\section{Introduction}

\vspace{-0.1in}
Neural networks (NN) are highly flexible and good approximators of complex functions at the expense of overparameterization, where the number of learnable parameters far exceeds the available training data. This increases computational demands and elevates the risk of overfitting \cite{Zhang2017generalize}, where the model starts to memorize noise rather than capture meaningful patterns. To avoid overconfidence and miscalibration, uncertainty quantification (UQ) in neural network predictions is crucial, especially in fields involving critical decision-making, such as clinical diagnostics or autonomous driving \cite{amodei2016concrete}. 

Addressing these challenges, Bayesian modeling presents a principled way to quantify the model prediction uncertainty~\cite{ghahramani2015probabilistic}. Moreover, the Bayesian framework demonstrates heightened resilience against noise and adversarial perturbations due to their inherent probabilistic predictive capabilities \cite{wicker2021bayesian}. To this effect, the confluence of deep neural networks and Bayesian inference in the form of Bayesian neural networks (BNN) has significantly advanced probabilistic machine learning. However, exact posterior inference is intractable in neural networks due to the extremely high dimensionality. Instead, approximation methods, such as mean-field variational inference \cite{Jordan_Graph-2000}, provide a computationally feasible way to perform posterior inference in BNNs. Nonetheless, this approximation severely limits the expressiveness of the inferred posterior, ultimately degrading the quality of the uncertainty estimates \cite{izmailov21bnn}. 

To this end, Bayesian inference over a low-dimensional subspace of neural network weights offers an elegant solution for improving accuracy, robustness, and uncertainty quantification \cite{izmailov2020subspace}. Alternatively, active subspace methods \cite{constantine2015active} developed for inverse problems involving computer models perform dimension reduction by constructing a linear subspace of inputs with directions accounting for most variation in the function's output. In this paper, we demonstrate the utility of active subspace inference for Bayesian deep learning. Accordingly, we identify a low-dimensional subspace embedded in a high-dimensional parameter space to capture most of the variability in the neural network output. Our main contributions:
\begin{itemize}[noitemsep,nolistsep]
    \item We propose two active subspace methods: output-informed (\textbf{AS}) and likelihood-informed (\textbf{LIS}) for scalable Bayesian inference in deep learning. The compact subspace facilitates the use of otherwise intractable posterior approximation methods.
    \item We empirically demonstrate the superior uncertainty quantification offered by our approach compared to Bayesian inference over the full network and an existing subspace inference method.
\end{itemize}
\vspace{1mm}

\noindent\textbf{Related work.}
The framework for interpretable inference through \textit{effective dimensionality} of the parameter space was provided by \cite{mackay1991bayesian} in the context of BNNs. \cite{maddox2020rethinking} demonstrated that near local optima, numerous directions in parameter space have low impact on neural network predictions. \cite{maddox2019simple} performed inference over low-dimensional subspace spanned by the SGD iterates of NN weights. \cite{izmailov2020subspace} further substantiated that posterior inference with a compact subspace of the entire parameter space is as effective as with the full network.

Alternately, active subspace finds its roots in the computer model UQ literature \cite{constantine2014active}, where dimension reduction is performed on the high-dimensional input space using model outputs. The identified active subspace is then exploited for cheap approximate modeling of the computationally expensive simulators \cite{constantine2015exploiting,loudon2017hiv}. The combination of active subspace methods and deep neural networks is still an underexplored area. Recently, \cite{cui2020active} employed active subspace to perform the neuron pruning at an intermediate layer, whereas \cite{tripathy2019deep} introduced a deep active subspace method to project high-dimensional neural network input to a compact subspace. However, none of these approaches reduce the parameter space dimensionality in NNs. Our proposed approach bridges this gap by using active subspaces for UQ in BNNs.
\vspace{-0.1in}
\section{Preliminaries}

\vspace{-0.1in}

\noindent{\textbf{Notations}}: We have bold lowercase letters $(\boldx)$ and bold uppercase letters $(\boldX)$ denoting vectors and matrices, respectively.

\vspace{-0.1in}
\subsection{Active subspace}
Active subspace dimension reduction seeks to identify the directions in variable space that have the most influence on the function's output on average. Consider a continuous function $f_{\btheta}(\boldx)$ with $\boldx$ and $\btheta$ denoting the input and the model parameters. In this work, we focus on the active directions of the parameter space that explain the most variance in the function's gradient, $\nabla_{\btheta} f_{\btheta}(\boldx)$, via eigendecomposition of
\begin{align}
\calC & = \E_{\btheta}\left[(\nabla_{\btheta} f_{\btheta}(\boldx)) (\nabla_{\btheta} f_{\btheta}(\boldx))^T\right] \nonumber \\
& = \int (\nabla_{\btheta} f_{\btheta}(\boldx)) (\nabla_{\btheta} f_{\btheta}(\boldx))^T p(\btheta) \boldd \boldx
\label{e:grad-mat}
\end{align}
where $\calC$ is an uncentered covariance matrix of the gradients which is symmetric semi-definite and admits the following eigendecomposition,
$$\calC = \boldV \bLambda \boldV^T = [\boldV_1 \enskip \boldV_2] \begin{bmatrix} \bLambda_1 & 0 \\ 0 & \bLambda_2 \end{bmatrix} [\boldV_1 \enskip \boldV_2]^T$$
where $\boldV$, $\bLambda$ contain the eigenvectors and eigenvalues respectively. Note, $\bLambda_1 = \rm{diag}(\lambda_1, \dots, \lambda_d)$ and $\bLambda_2 = \rm{diag}(\lambda_{d+1}, \dots, \lambda_n)$ with $\lambda_1 \ge \dots \lambda_n \ge 0 $. The subspace spanned by $\boldV_1$ corresponds with the $d$ largest eigenvalues in $\bLambda_1$ and is considered as the ``active'' subspace. Accordingly, $f_{\btheta}(\boldx)$ is most sensitive to random perturbations in $\boldV_1$.

\vspace{-0.1in}
\subsection{Bayesian model formulation} 
Let $\mathcal{D}=\{(\boldx_i,y_i)\}_{i=1,\cdots,N}$ denote a training dataset of $N$ i.i.d. observations where $\boldx$ represents inputs and $y$ denotes corresponding outputs. In the Bayesian framework, instead of optimizing over a single probabilistic model, $p(y|\boldx,\btheta)$, we discover all likely models via posterior inference over model parameters. The Bayes' rule provides the posterior distribution: $p(\btheta|\mathcal{D}) \propto p(\mathcal{D}|\btheta) p(\btheta)$, where $p(\mathcal{D}|\btheta)$ denotes the likelihood of $\mathcal{D}$ given the model parameters $\btheta$ and $p(\btheta)$ is the prior distribution over the parameters. Given $p(\btheta|\mathcal{D})$ we predict the label corresponding to new example $\boldx^*$ by Bayesian model averaging (BMA) using Monte Carlo sampling:
\begin{align*}
p(y^*|\boldx^*,\calD) &= \int p(y^*|\boldx^*,\btheta) p(\btheta|\calD) d \btheta \\
& \approx \frac{1}{M} \sum_{m=1}^M p(y^*|\boldx^*,\btheta_m) , \enskip \enskip \btheta_m \sim p(\btheta|\calD)
\end{align*}

\vspace{-0.1in}
\subsection{Approximate Bayesian inference} 
Markov chain Monte Carlo (MCMC) sampling is the gold standard for Bayesian model inference. In particular, we can use exact full-batch Hamiltonian Monte Carlo (HMC) to approximately sample from $p(\btheta|\calD)$. Alternatively, variational inference (VI) leverages deterministic optimization to speed up the inference \cite{Blei2017}. It infers a variational distribution on the model parameters $q(\btheta)$ by minimizing the Kullback-Leibler (KL) divergence from the true Bayesian posterior $p(\btheta|\mathcal{D})$:
$$ \widehat{q}(\btheta)=\underset{q(\btheta) \in \mathcal{Q}}{\text{argmin}}\:\: d_{\rm KL}(q(\btheta),p(\btheta|\mathcal{D}))$$
where $\mathcal{Q}$ denotes a family of variational distributions. The above optimization problem to infer the variational distribution parameters is typically solved by minimizing the negative evidence lower bound (ELBO), which is defined as
\begin{equation}
\label{e:elbo} 
    \mathcal{L}(\btheta)= -\mathbb{E}_{q(\btheta)} [\log p(\mathcal{D}|\btheta)]+d_{\rm KL}(q(\btheta),p(\btheta)),
\end{equation}
where the first term is the data-dependent cost widely known as the negative log-likelihood (NLL), and the second is prior-dependent and serves as regularization. Stochastic gradient descent algorithms can be derived to optimize \eqref{e:elbo}  \cite{Kingma-welling-2014}. 

\vspace{-0.1in}
\section{Active Subspace Learning}

\vspace{-0.1in}
A common approach for dimension reduction is principal component analysis (PCA), which can be used to identify a low-rank linear subspace of parameters $\btheta$ that explains most of the variance in the prior $\pi(\btheta)$. 
However, PCA does not consider any information about a function of the parameters $f(\btheta)$, such as a neural network. Active subspace (AS) reduction seeks to identify the parameter directions that have the greatest influence on the neural network's output. 

For $\boldx \in \mathbb{R}^p$, consider a neural network with $L$ hidden layers. The weight matrix and bias vector in the $l^{\text{th}}$ layer are denoted by $\boldW_l$ and $\boldb_l$, and $\psi(.)$ is the activation function. With $\btheta=\{\boldW_1,\boldb_1, \cdots, \boldW_L,\boldb_L\}$, the NN output is:
$$\eta_{\btheta}(\boldx)=\boldb_{L}+\boldW_{L}\psi(\boldb_{L-1}+\boldW_{L-1} \psi( \cdots \psi(\boldb_{1}+\boldW_{1}\boldx)))$$

If $\eta_{\btheta}(\boldx)$ is a univariate outcome, we set $f_{\btheta}(\boldx)=\eta_{\btheta}(\boldx)$ in \eqref{e:grad-mat}. We call this approach the outcome-informed active subspace (\textbf{AS}) method. Alternatively, we employ a generalization of the \textbf{AS} using optimal ridge function approximations \cite{baptista2022gradient,oleary2022derivative,Zahm2020}, especially in multivariate output settings. In particular, we use the mean squared error (MSE) loss ($\calL_{\btheta}(\boldx)$) for active subspace construction by setting $f_{\btheta}(\boldx)=\calL_{\btheta}(\boldx)$ in \eqref{e:grad-mat}. We term this approach likelihood-informed active subspace (\textbf{LIS}) since MSE loss is nothing but the negative log-likelihood of Gaussian distribution. LIS has been previously explored for computer model UQ \cite{Zahm2020, Cui2014, Cui2022}
and follows a similar philosophy to AS, where the function to analyze sensitivities is now the log-likelihood given data. 


Let $\calS$ be our proposed K-dimensional subspace defined as:
$$\calS = \{\btheta|\btheta = \hat{\btheta}_0 + \boldP \boldz = \hat{\btheta}_0 + z_1\hat{\boldv}_1+\cdots+z_K\hat{\boldv}_K\},$$
where $\hat{\btheta}_0$ are the pretrained weights, $\boldP = (\hat{\boldv}_1^T,\cdots,\hat{\boldv}_K^T)$ is the projection matrix, and $\boldz = (z_1,\cdots,z_K)$ are the $K$ subspace parameters on which we perform Bayesian inference. We use pretrained SWAG model weights as $\hat{\btheta}_0$ for a fair comparison with the SGD-PCA subspace method \cite{izmailov2020subspace}. We choose sufficiently diffused Gaussian prior $\calN(0,\tilde{\sigma}^2 \boldI_K)$ on $\boldz$ parameters.

\begin{algorithm}[H]
\caption{Active subspace construction and inference} \label{alg:AS}
\begin{algorithmic}[1]
\State Input: NN output or log-likelihood function $f$, pretrained weights $\hat{\btheta}_0$, \# of gradient samples $M$, active subspace dimension $K$, projection matrix for subspace $P$, perturbation standard deviation $\sigma_0$, \# of BMA samples J.
\For {$m= 1,2,\dots, M$}
\State Sample $\boldx_m, y_m \in \boldx, y$ 
\State Sample $\btheta_m \sim \calN(\hat{\btheta}_0,\sigma_0^2\boldI)$ 
\State Compute gradients: $\nabla_{\btheta_m} f(\boldx_m)$
\EndFor
\State $\hat{\calC} = \frac{1}{M} \sum_{m=1}^M (\nabla_{\btheta_m} f(\boldx_m)) (\nabla_{\btheta_m} f(\boldx_m))^T$
\State SVD decomposition: $\hat{\calC} = \hat{\boldV} \hat{\bLambda} \hat{\boldV}^T$
\State Projection matrix: $ \boldP = \hat{\boldV}^T$
\State Posterior inference over subspace parameters $\boldz$
\State $\hat{\btheta} = \hat{\btheta}_0 + \frac{1}{J} \sum_{j=1}^J \boldP \hat{\boldz}_j, \quad \hat{\boldz} \sim p(\boldz|\calD)$
\end{algorithmic}
\end{algorithm} 
\vspace{-0.1in}
\section{Experiments}

\vspace{-0.1in}
In this section, we demonstrate the performance of our proposed \textbf{AS} and \textbf{LIS} subspace inference approaches on several regression tasks. We consider multilayer perceptron (MLP) and implement it in PyTorch \cite{PyTorch2019NeurIPS}.
\vspace{1mm}

\noindent\textbf{Baselines.} Our baselines include the deterministic NN (trained with SGD), full BNN, and SGD-PCA subspace method. We perform Bayesian inference using No-U-Turn-Sampler in the simulation study and VI \cite{blundell2015weight} in UCI regression datasets.
\vspace{1mm}

\noindent\textbf{Metrics.} We use posterior model predictions for qualitative assessment of uncertainty estimates in the simulation study. We assess the performance using root mean squared error (RMSE), log-likelihood, and calibration to 95\% credible intervals on 
test data in UCI regression tasks.

\vspace{-0.1in}
\subsection{Uncertainty quantification in univariate regression}
\begin{figure*}[t]
\begin{subfigure}[b]{0.195\textwidth}
    \centering
    \includegraphics[width=\textwidth]{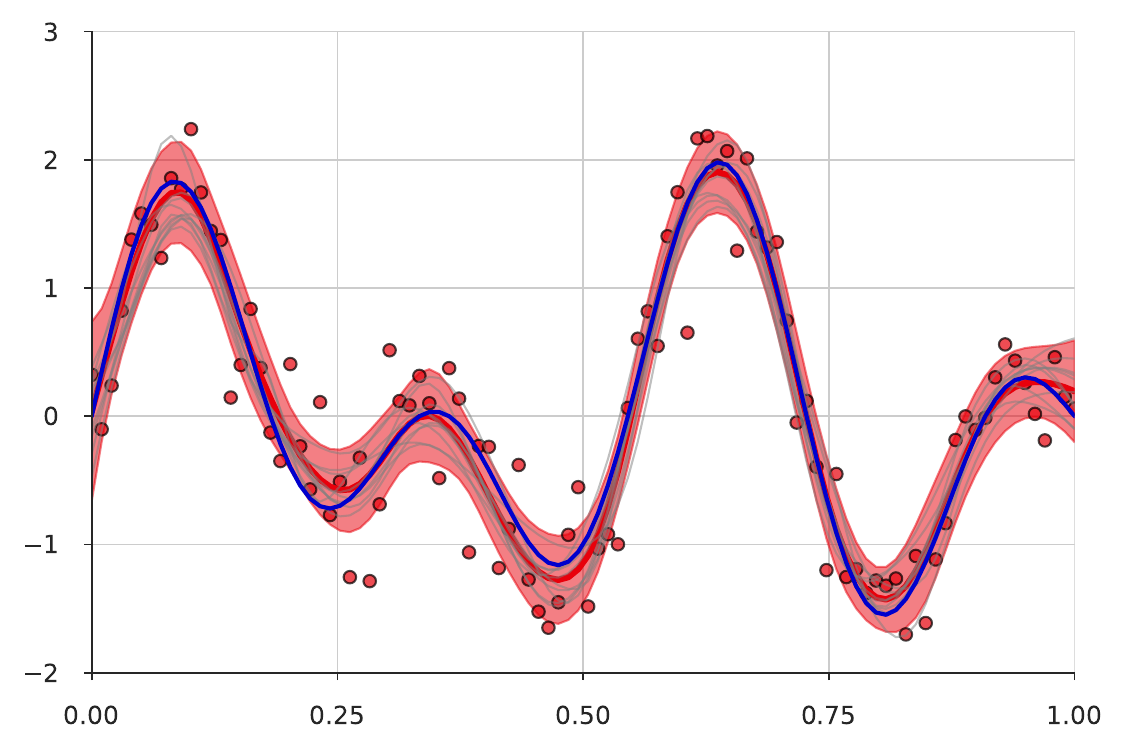}
\end{subfigure}
\hfill
\begin{subfigure}[b]{0.195\textwidth}  
    \centering 
    \includegraphics[width=\textwidth]{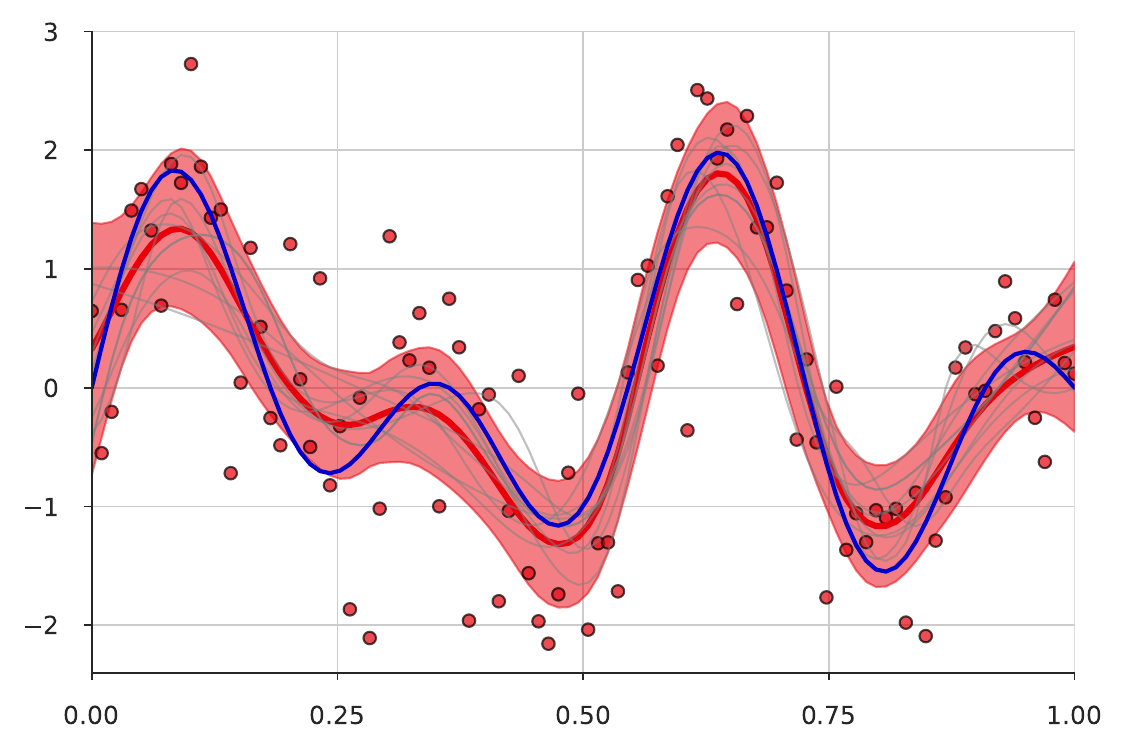}
\end{subfigure}
\hfill
\begin{subfigure}[b]{0.195\textwidth}  
    \centering 
    \includegraphics[width=\textwidth]{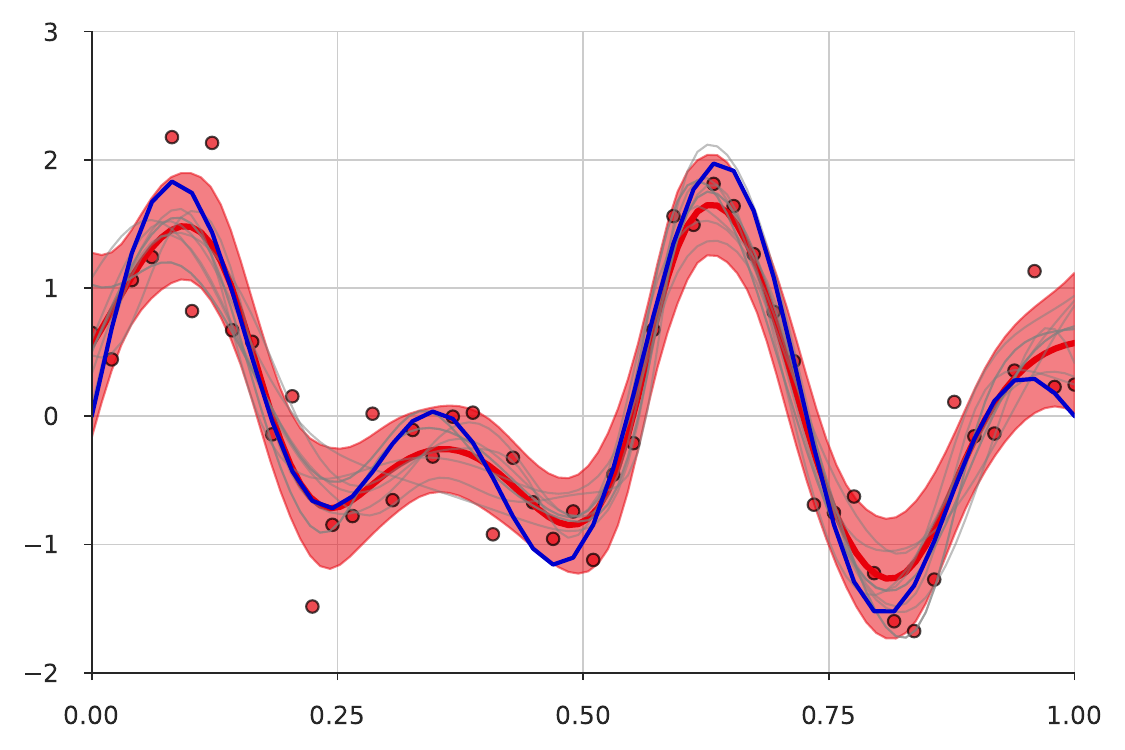}
\end{subfigure}
\hfill
\begin{subfigure}[b]{0.195\textwidth}  
    \centering 
    \includegraphics[width=\textwidth]{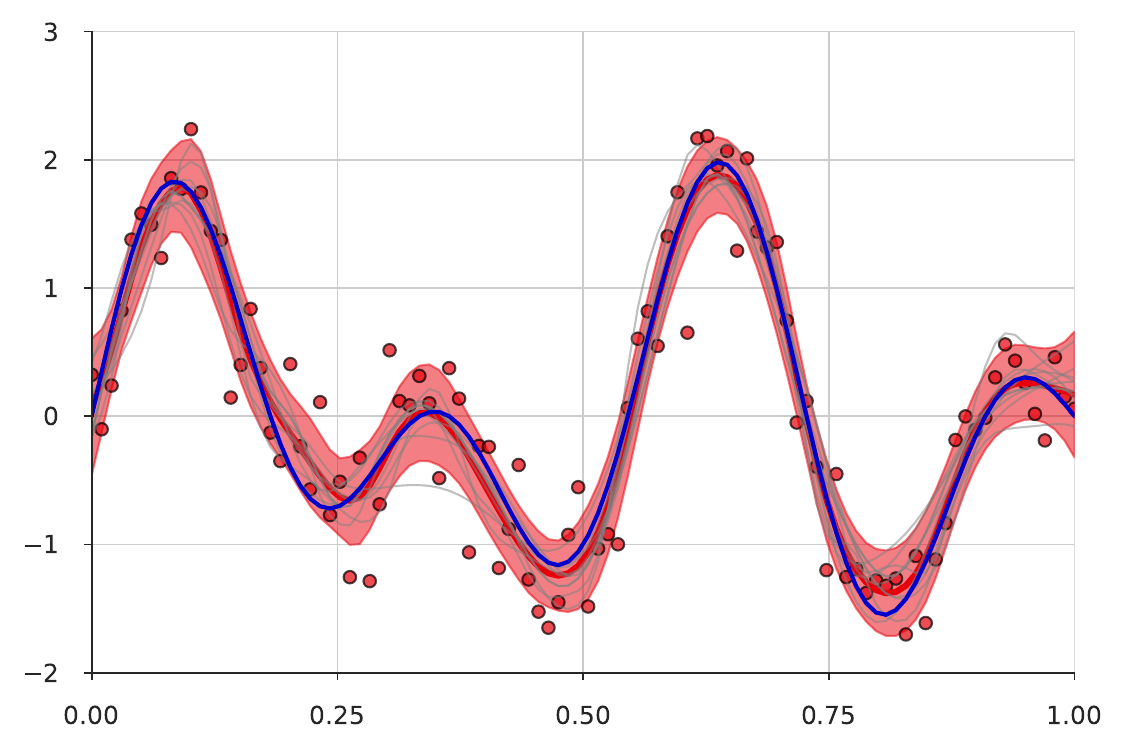}
\end{subfigure}
\hfill
\begin{subfigure}[b]{0.195\textwidth}  
    \centering 
    \includegraphics[width=\textwidth]{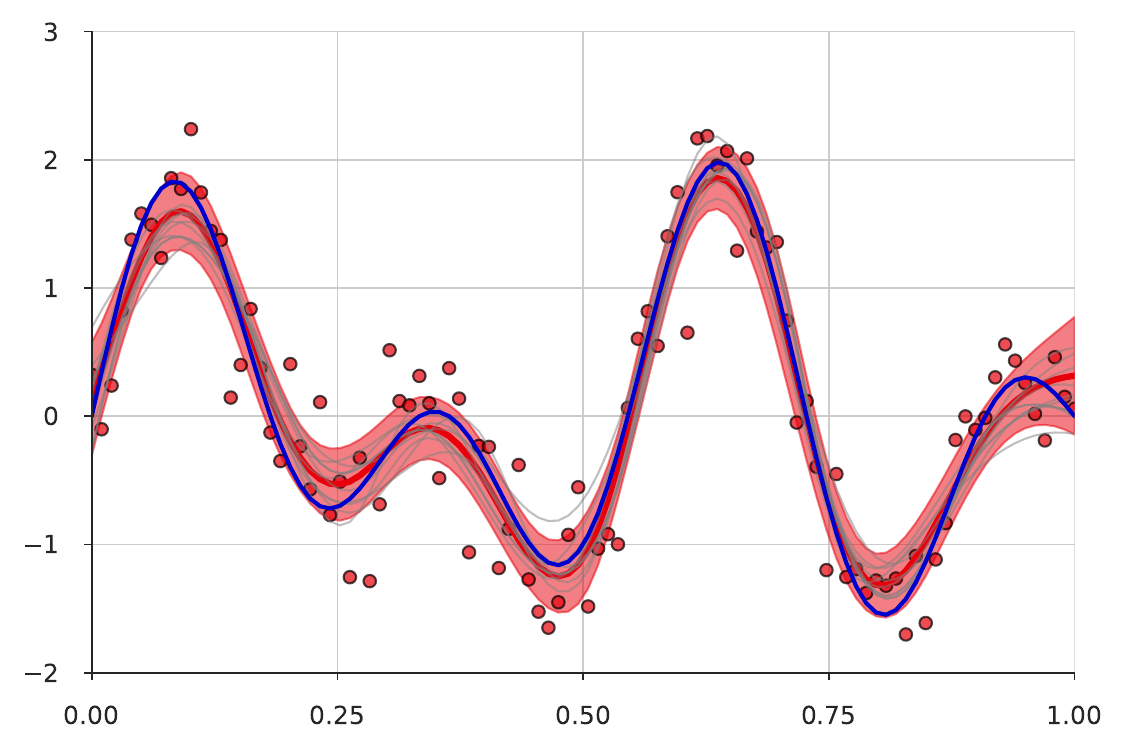}
\end{subfigure}
\begin{subfigure}[b]{0.195\textwidth}
    \centering
    \includegraphics[width=\textwidth]{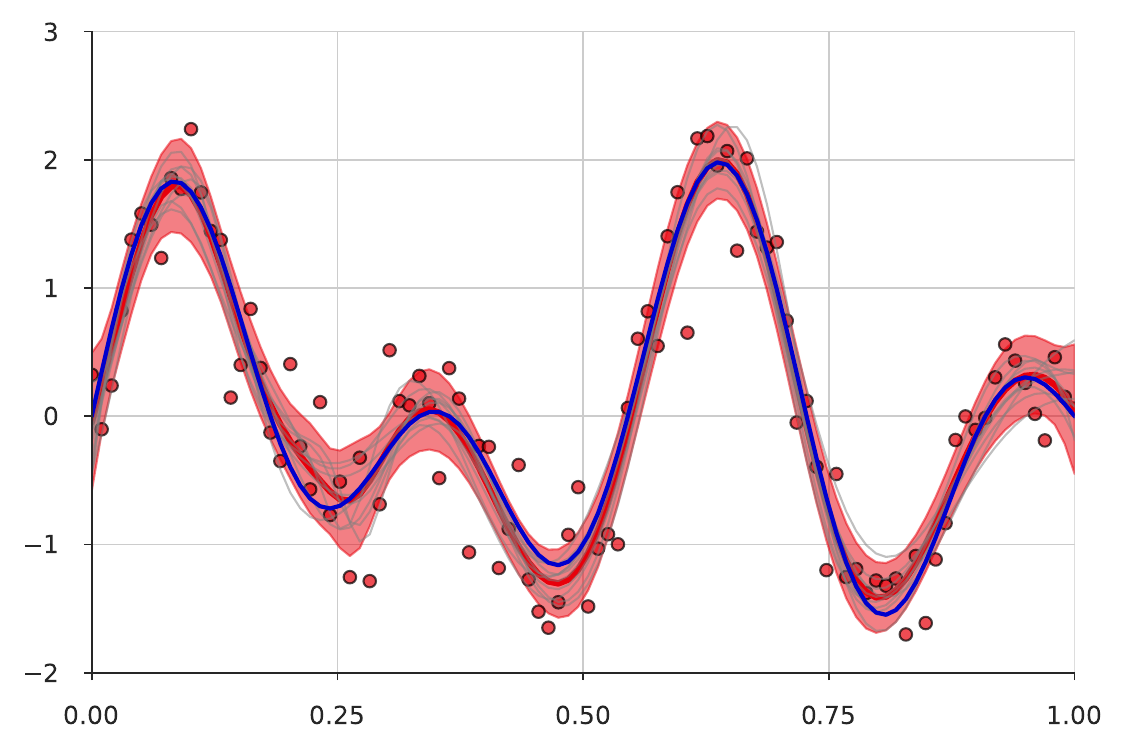}
\end{subfigure}
\hfill
\begin{subfigure}[b]{0.195\textwidth}  
    \centering 
    \includegraphics[width=\textwidth]{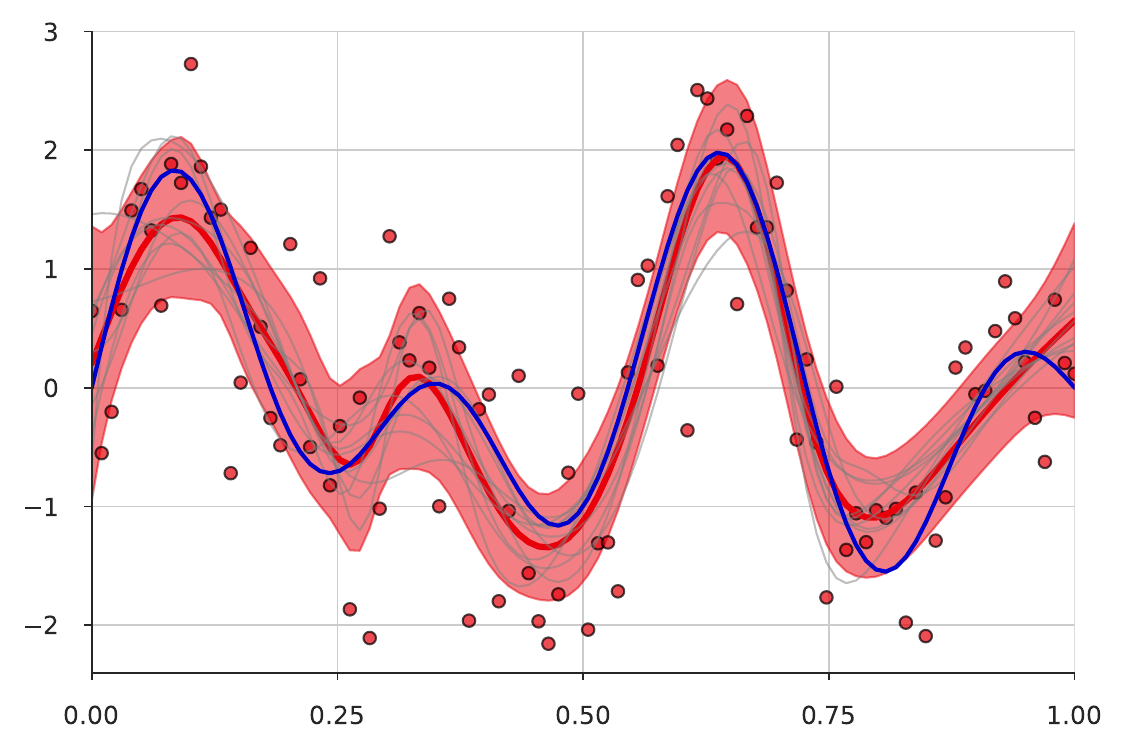}
\end{subfigure}
\hfill
\begin{subfigure}[b]{0.195\textwidth}  
    \centering 
    \includegraphics[width=\textwidth]{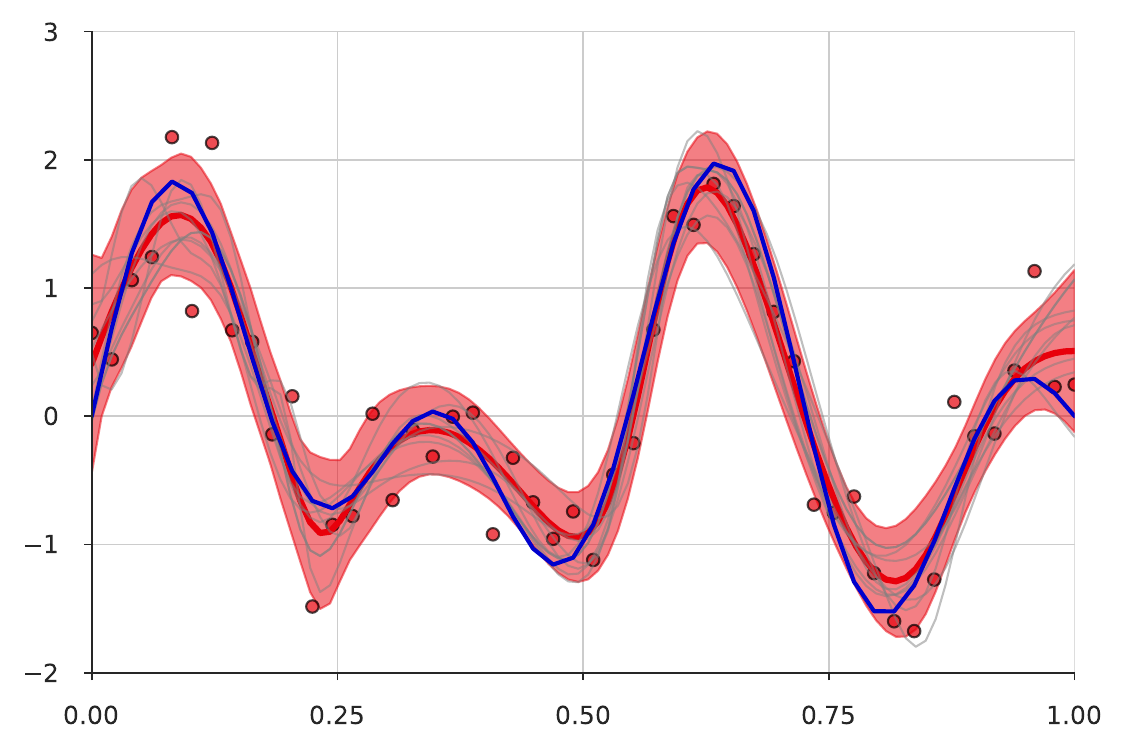}
\end{subfigure}
\hfill
\begin{subfigure}[b]{0.195\textwidth}  
    \centering 
    \includegraphics[width=\textwidth]{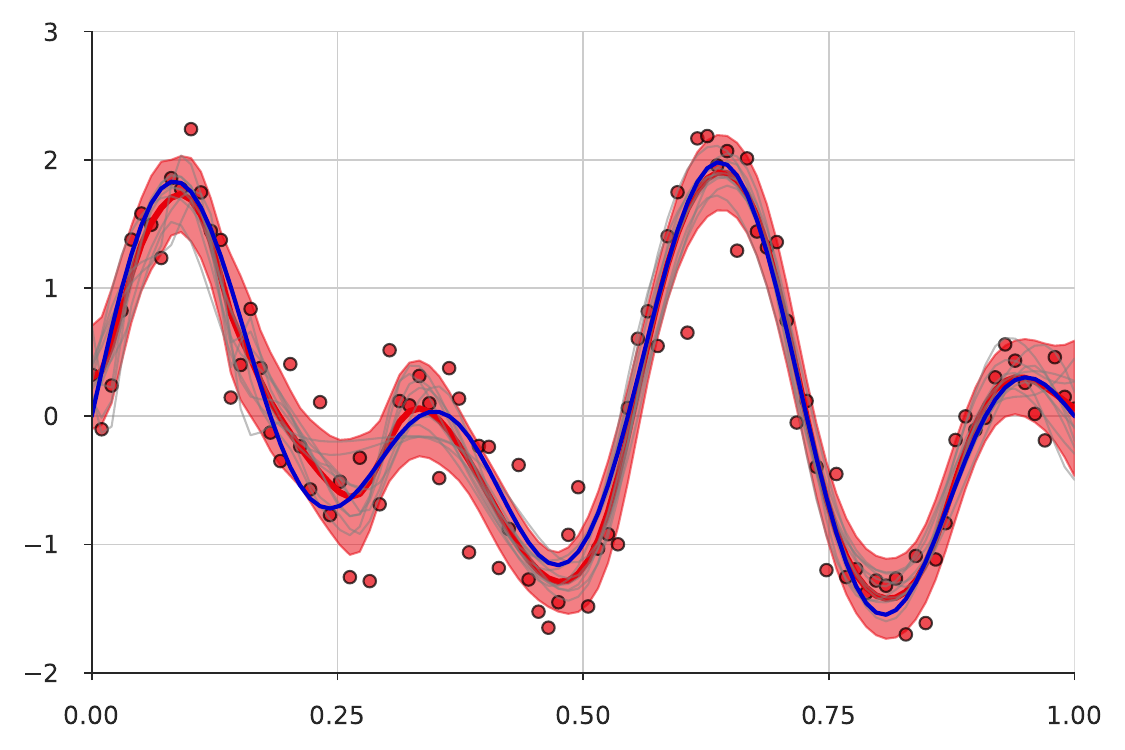}
\end{subfigure}
\hfill
\begin{subfigure}[b]{0.195\textwidth}  
    \centering 
    \includegraphics[width=\textwidth]{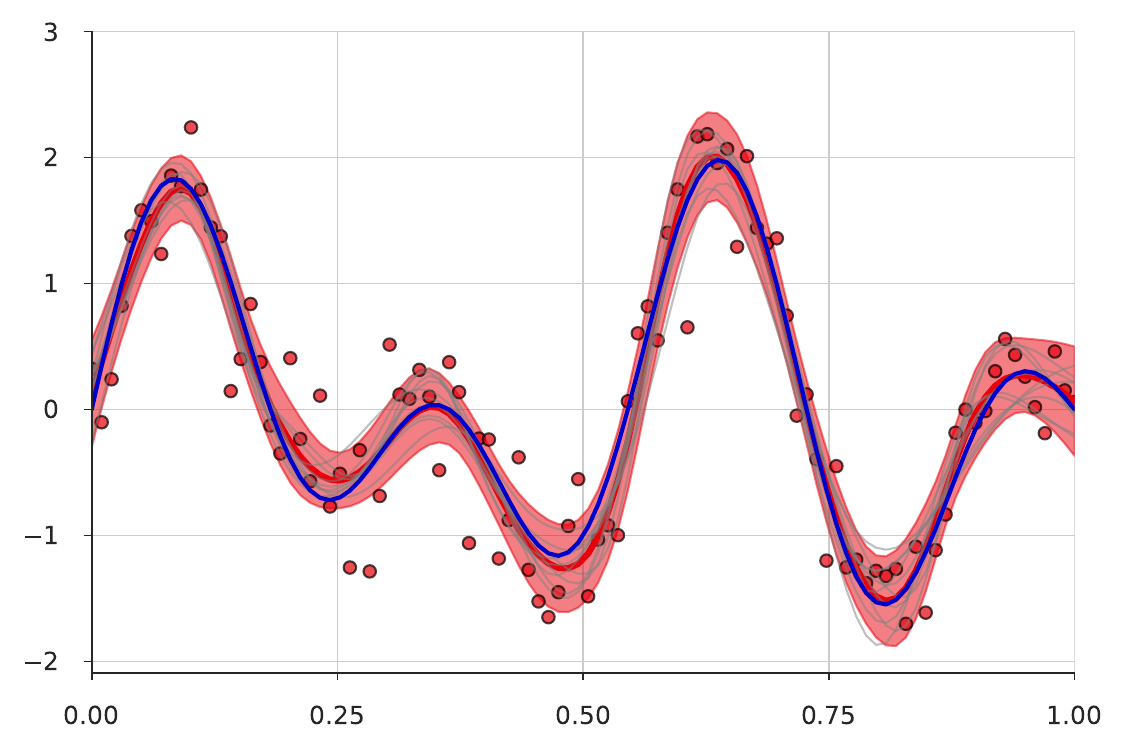}
\end{subfigure}
\begin{subfigure}[b]{0.195\textwidth}
    \centering
    \includegraphics[width=\textwidth]{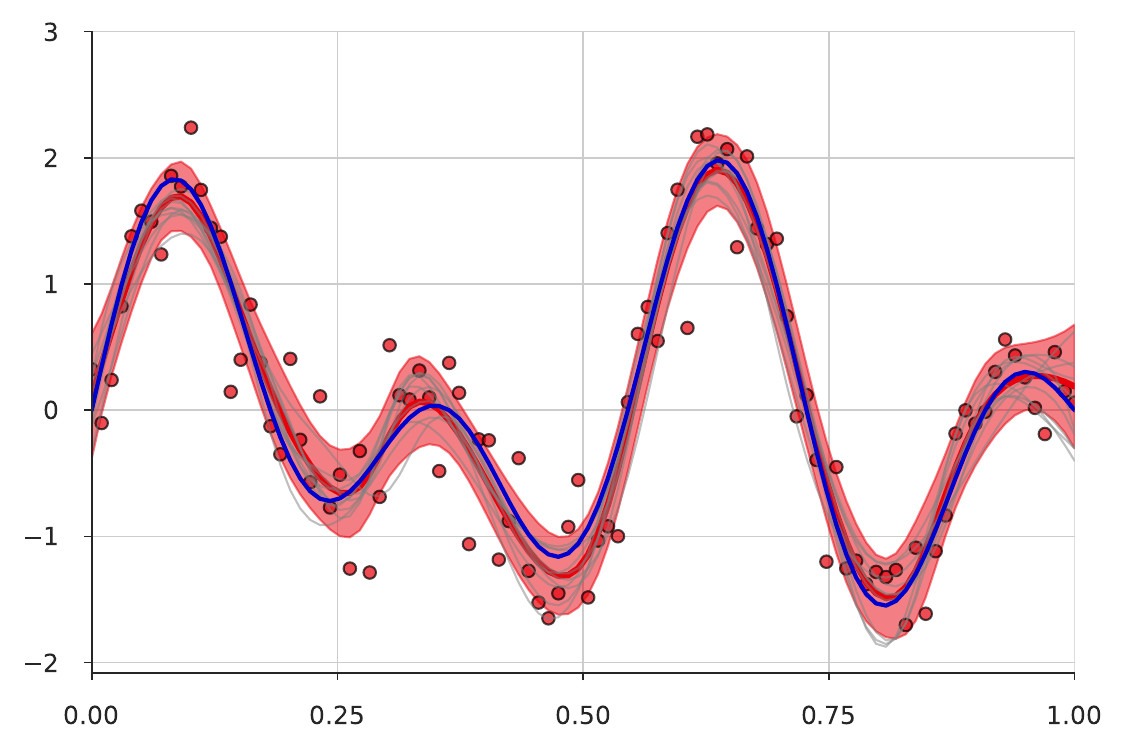}
\end{subfigure}
\hfill
\begin{subfigure}[b]{0.195\textwidth}  
    \includegraphics[width=\textwidth]{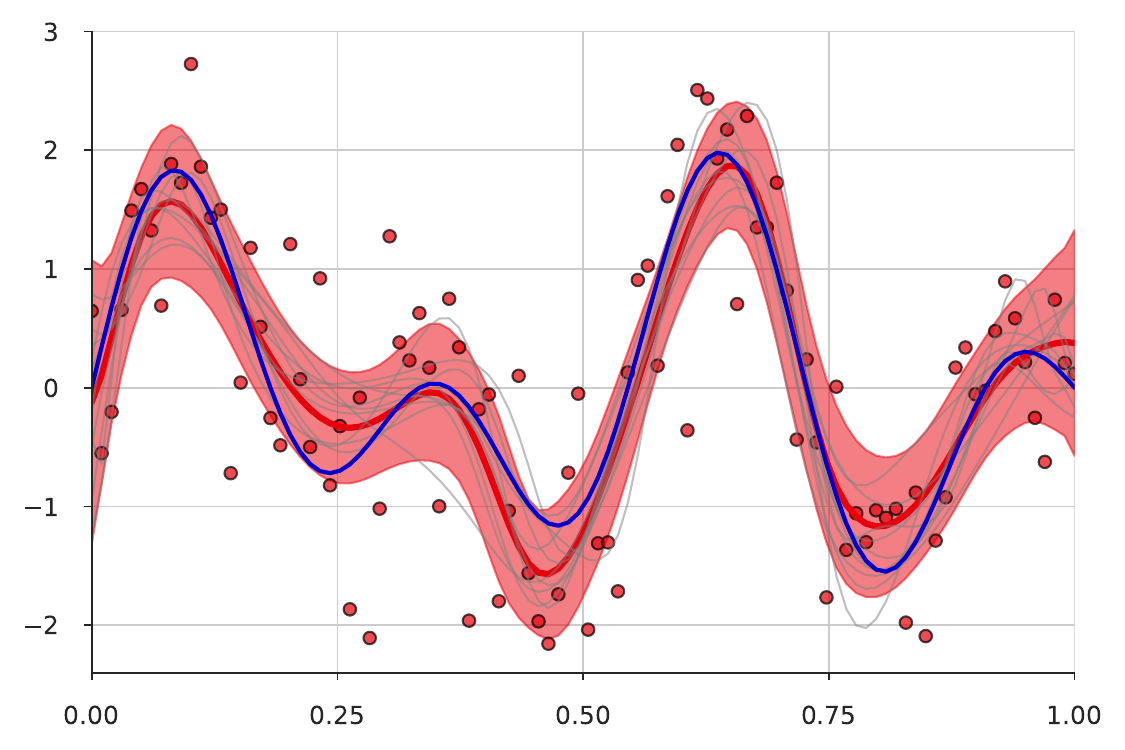}
\end{subfigure}
\hfill
\begin{subfigure}[b]{0.195\textwidth}
    \centering
    \includegraphics[width=\textwidth]{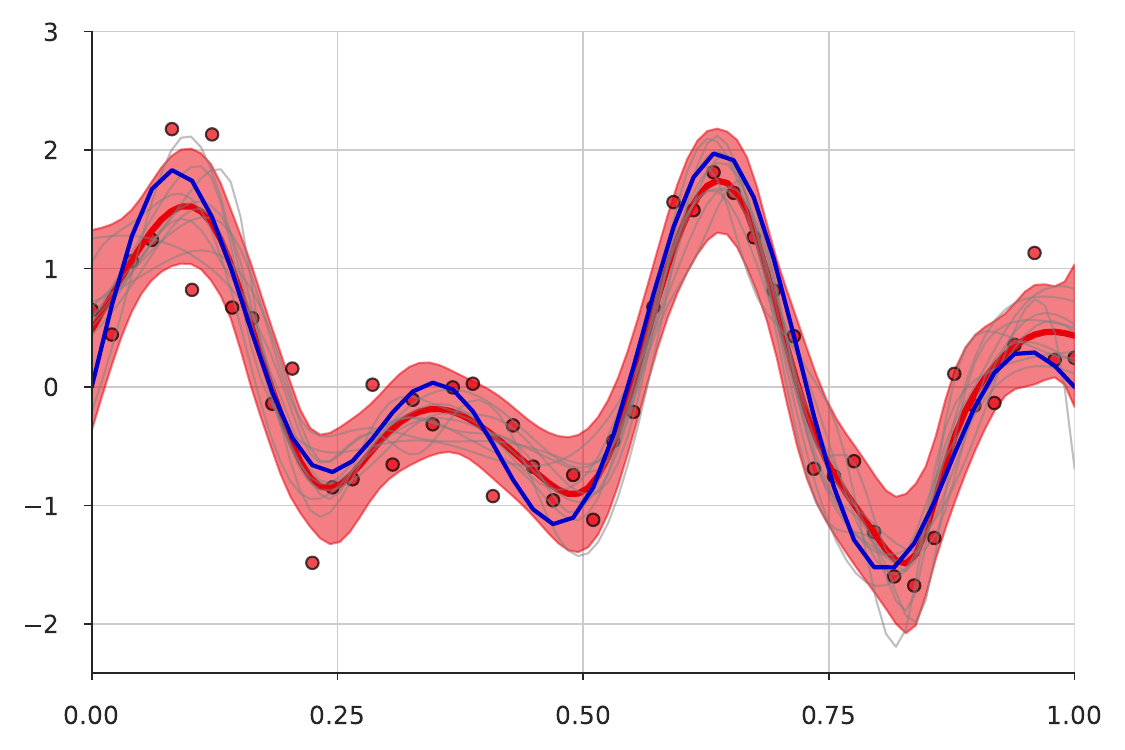}
\end{subfigure}
\hfill
\begin{subfigure}[b]{0.195\textwidth}  
    \centering 
    \includegraphics[width=\textwidth]{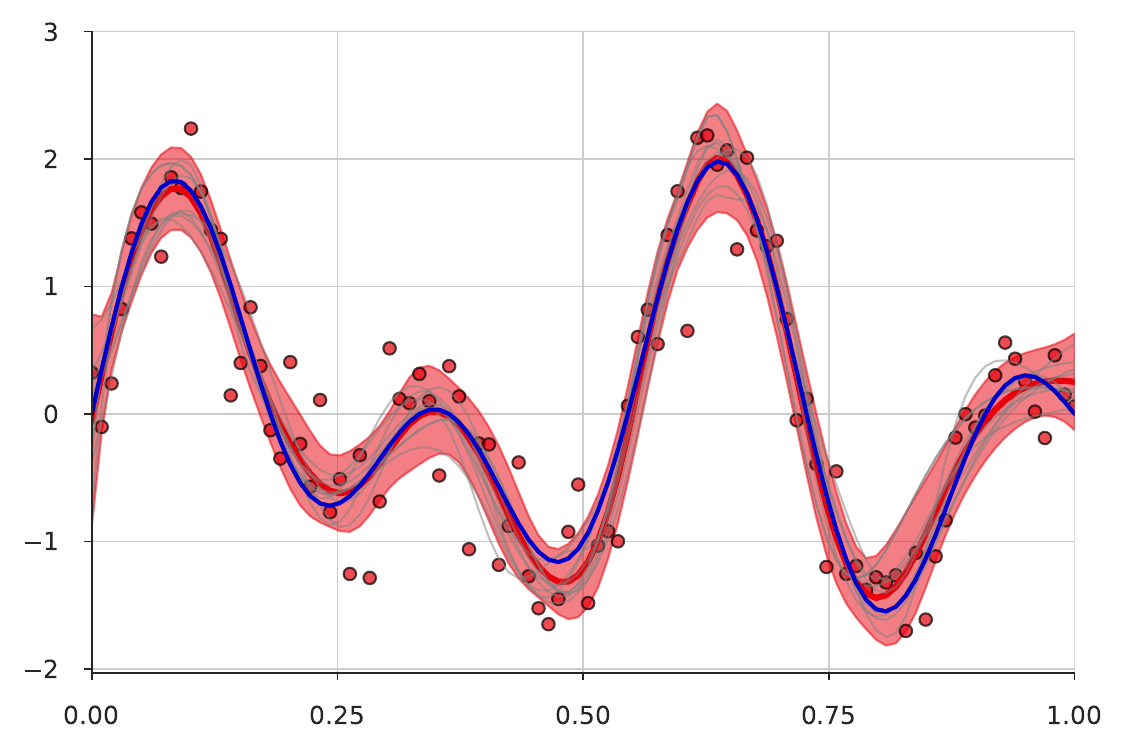}
\end{subfigure}
\hfill
\begin{subfigure}[b]{0.195\textwidth}  
    \centering 
    \includegraphics[width=\textwidth]{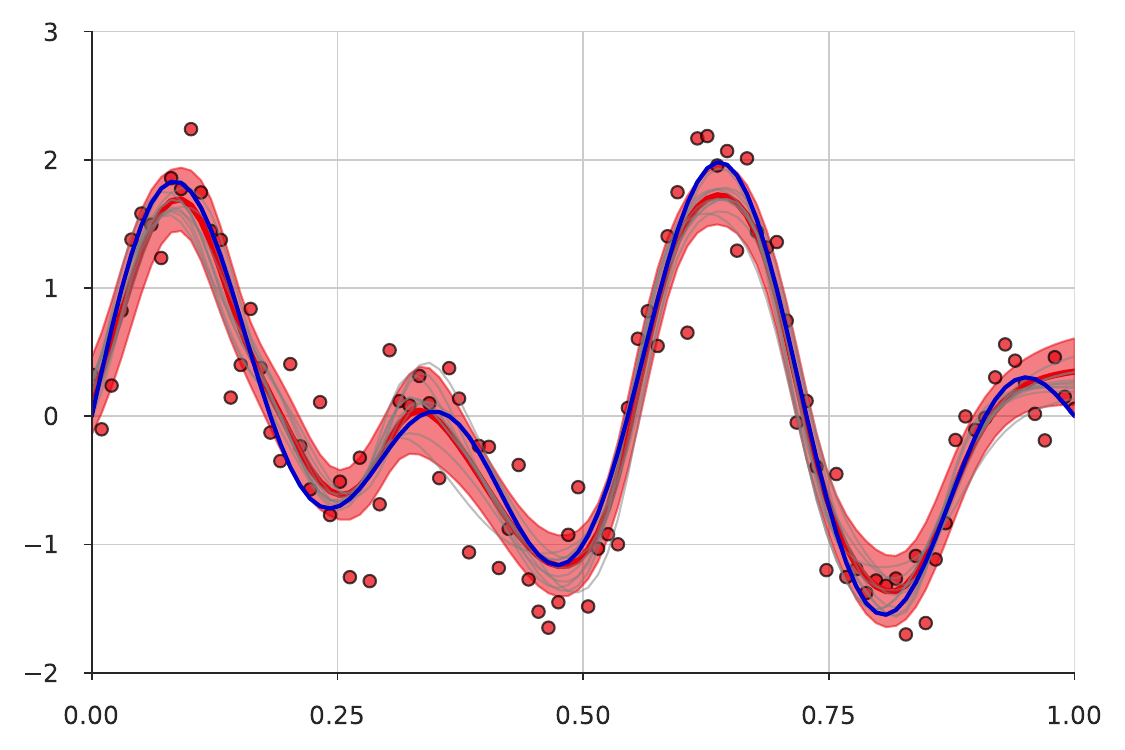}
\end{subfigure}
\begin{subfigure}[b]{0.195\textwidth}
    \centering
    \includegraphics[width=\textwidth]{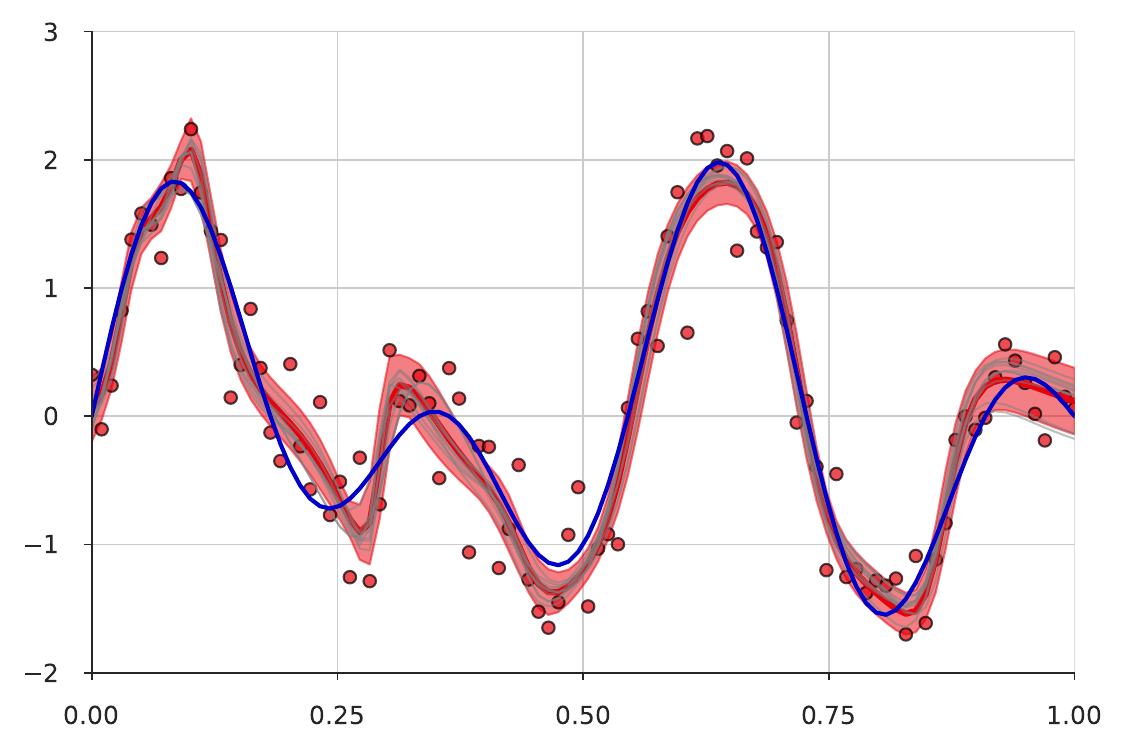}
\end{subfigure}
\hfill
\begin{subfigure}[b]{0.195\textwidth}  
    \centering 
    \includegraphics[width=\textwidth]{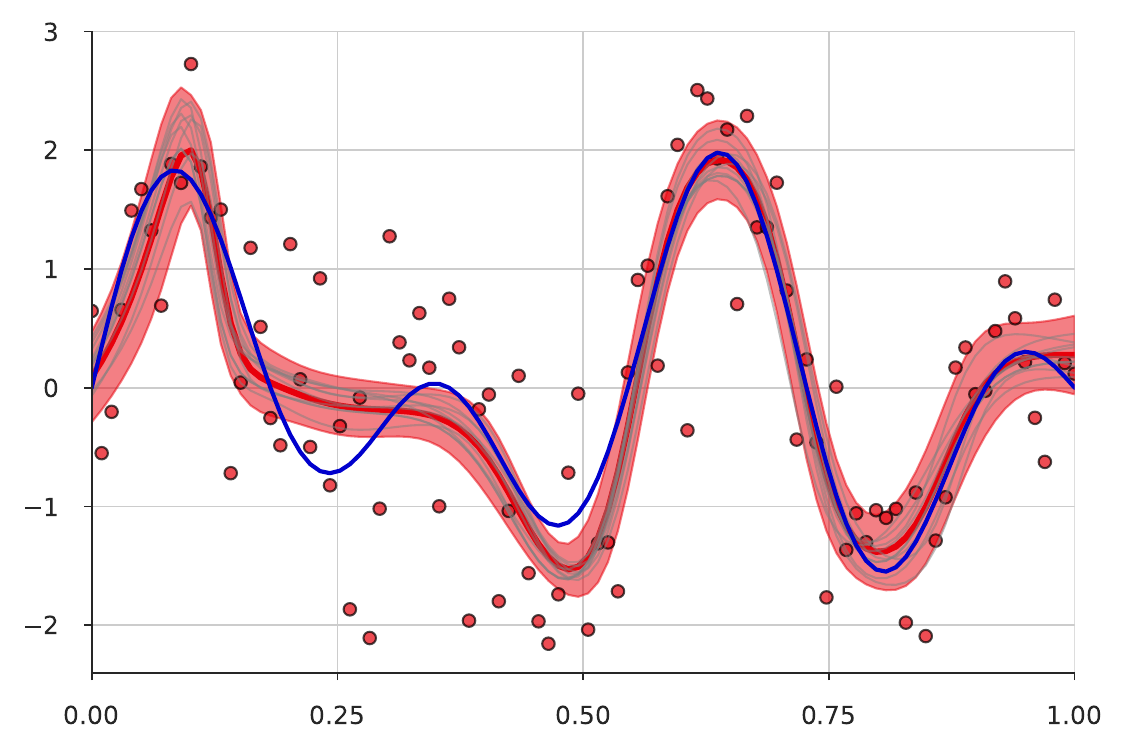}
\end{subfigure}
\hfill
\begin{subfigure}[b]{0.195\textwidth}  
    \centering 
    \includegraphics[width=\textwidth]{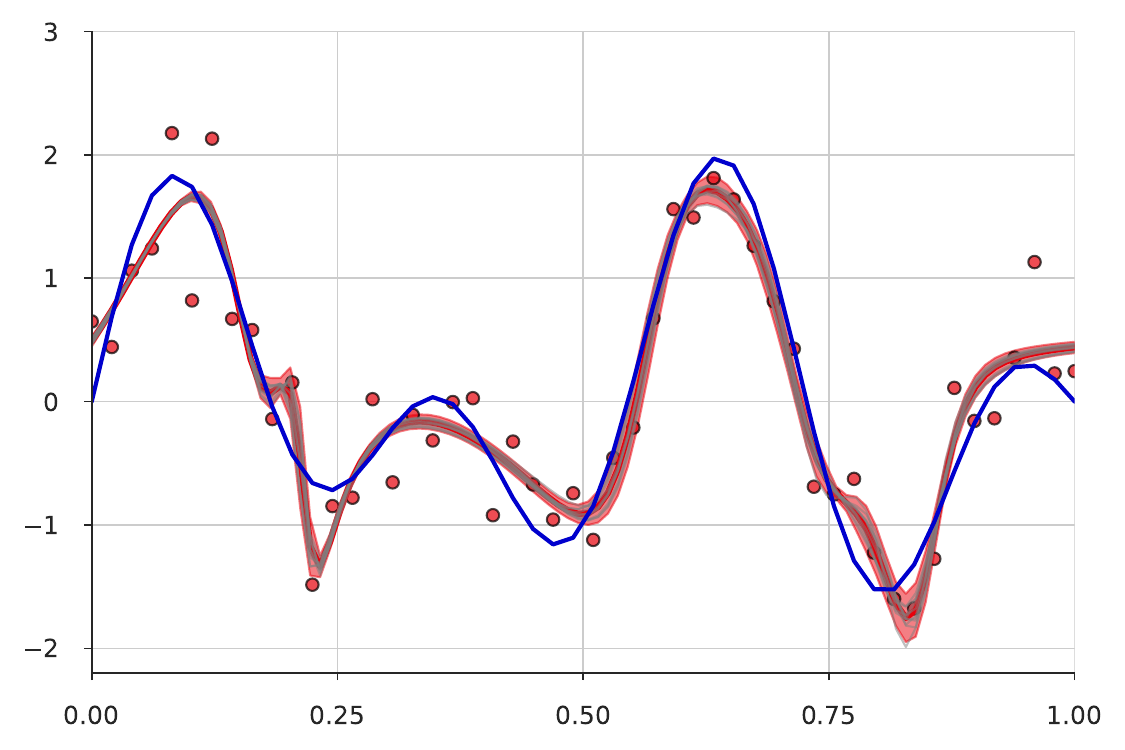}
\end{subfigure}
\hfill
\begin{subfigure}[b]{0.195\textwidth}  
    \centering 
    \includegraphics[width=\textwidth]{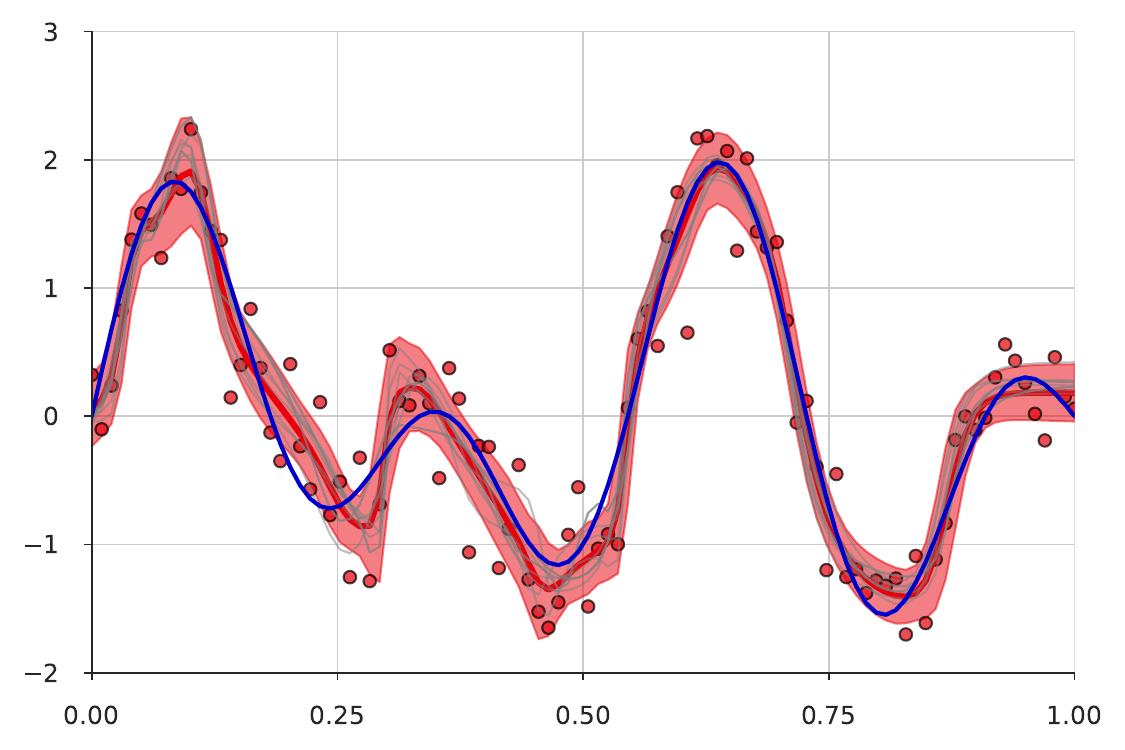}
\end{subfigure}
\hfill
\begin{subfigure}[b]{0.195\textwidth}  
    \centering 
    \includegraphics[width=\textwidth]{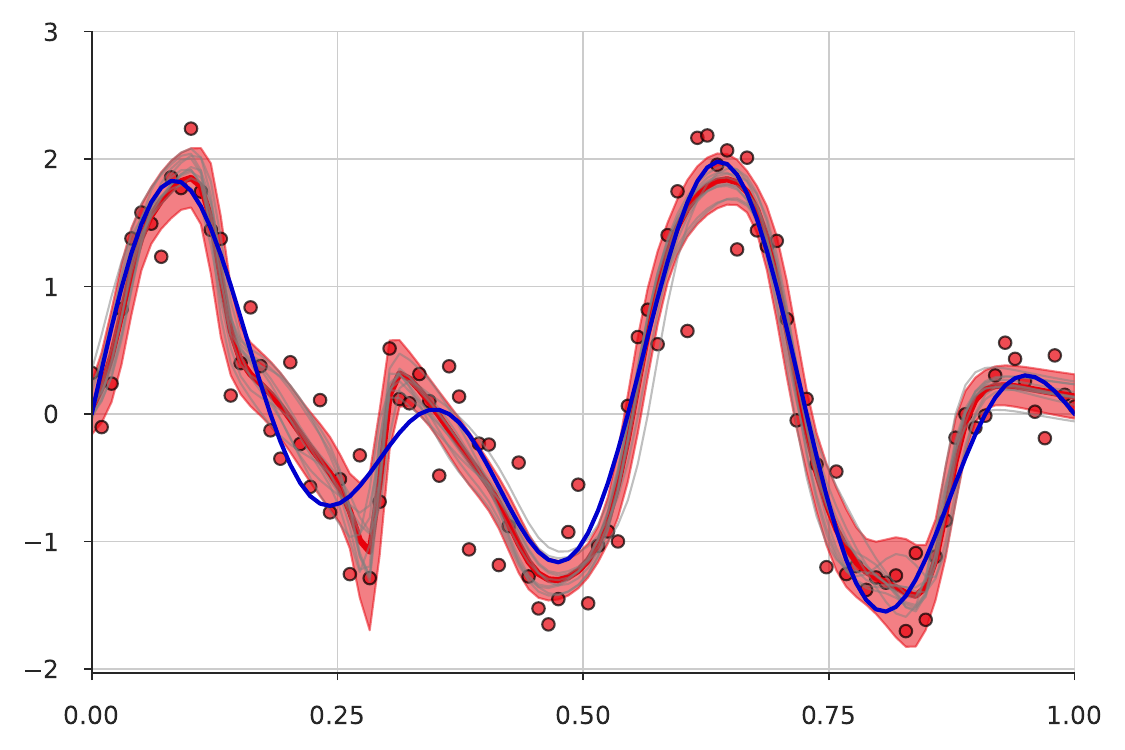}
\end{subfigure}
\caption{Posterior model predictions inferred from 4 models in 5 scenarios. Data (red dots), true function (blue line), predictive mean (red line), sampled functional posteriors (gray lines), $\pm$2 standard deviations from mean (shaded area). \textbf{Rows}: (1) full BNN (2) AS (3) LIS  (4) SGD-PCA models. \textbf{Columns}: (1) $(\sigma_{\epsilon}=0.4, N=100)$, (2) $(\sigma_{\epsilon}=0.8, N=100)$, (3) $(\sigma_{\epsilon}=0.4, N=50)$, (4) 6-layer network with 64 nodes each, (5) 3-layer network with 128 nodes each. The first 3 columns use a 3-layer network with 32 nodes each. $\sigma_{\epsilon}$ and $N$ are noise standard deviation and number of data points respectively. The last 2 columns use the same data as the first column. Our AS and LIS models provide uncertainty estimates comparable to the full BNN. SGD-PCA is overconfident and provides narrow uncertainty bands around the predictive mean, especially in the first 3 modeling scenarios}
\label{fig:nn-as}
\end{figure*}

We have studied neural network weight inference for multilayer perceptrons in a univariate regression setting in \autoref{fig:nn-as}. We consider 5 scenarios with different data generation conditions and network architectures. The data generating process: $y = \sin(4\pi x) + \sin(7\pi x) + \epsilon$ with $\epsilon \sim \calN(0,\sigma_{\epsilon}^2)$. In \autoref{fig:nn-as} column 1 uses $\sigma_{\epsilon}=0.4$ and $N=100$ data samples. Columns 2 and 3 use $\sigma_{\epsilon}=0.8, N=100$ and $\sigma_{\epsilon}=0.4, N=50$ respectively. The first 3 columns use a 3-layer network with 32 nodes in each layer. Columns 4 and 5 use $\sigma_{\epsilon}=0.4, N=100$ together with 6 layers of 64 nodes each and 3 layers of 128 nodes each respectively. The Tanh activation function is chosen. 
Row 1 is full BNN, rows 2 and 3  are our 20-dimensional AS and LIS models, and row 4 is a 20-dimensional SGD-PCA model. We use $M=100$ gradient samples to construct our subspace and $J=30$ for BMA during inference. For a fair comparison, we use $\hat{\btheta}_0 = \hat{\btheta}_{\text{swag}}$, same as SGD-PCA, and 100 weight deviations in their subspace construction.

In all the scenarios, AS and LIS models provide uncertainty estimates that closely match the full BNN model. On the other hand, the SGD-PCA subspace model is overconfident in its predictions leading to narrower uncertainty bands around its predictive mean. This highlights the superior UQ provided by our models in various scenarios.

\vspace{-0.1in}
\subsection{UCI regression datasets}
We apply our methods (trained with VI) to UCI regression datasets and contrast their performance against deterministic NN (trained with SGD), BNN, and SGD-PCA models both trained with VI. We report the results averaged over 20 trials while splitting data randomly at 9:1 train-test ratio in each trial. We take a single hidden layer NN with 50 neurons and ReLU activation. For a given input $\boldx$ we produce two outputs: predictive mean $\mu_{\btheta}(\boldx)$ and predictive variance $\sigma^2_{\btheta}(\boldx)$. In our subspace construction: AS uses $f_{\btheta}(\boldx)=\mu_{\btheta}(\boldx)$ and LIS uses $f_{\btheta}(\boldx)=(y-\mu_{\btheta}(\boldx))^2/\sigma^2_{\btheta}(\boldx)$. VI with Adam \cite{Kingma2015Adam} is implemented to train all Bayesian models. 
\begin{table}[b!]
\fontsize{7.2}{9}\selectfont
  \caption{Unnormalized test data log-likelihoods on UCI regression datasets for our AS and LIS models and baselines. SGD: deterministic NN trained with SGD, VI: BNN trained with VI, SGD-PCA: SGD-PCA model trained with VI. Bold numbers are the best log-likelihoods.}
  \label{table:uci_small_likelihoods}
  \centering
  \setlength\tabcolsep{4pt}
  \begin{tabular}{lHHccccc}
    \toprule
    dataset & N & D & SGD & VI & SGD-PCA & AS & LIS \\
    \midrule
    boston & 506 & 13 & -2.75$\pm$0.13 & \textbf{-2.72}$\pm$0.07 & -2.73$\pm$0.13 & -2.76$\pm$0.14 & \textbf{-2.72}$\pm$0.18 \\
    concrete & 1030 & 8 & -3.17$\pm$0.20 & -3.29$\pm$0.05 & -3.09$\pm$0.15 & -3.05$\pm$0.12 & \textbf{-3.04}$\pm$0.10 \\
    energy & 768 & 8 & -2.36$\pm$0.03 & \textbf{-2.18}$\pm$0.06 & -2.39$\pm$0.03 & -2.41$\pm$0.03 & -2.35$\pm$0.06 \\
    naval & 11934 & 14 & \hphantom{-}5.44$\pm$1.53 & \hphantom{-}5.21$\pm$0.11 & \hphantom{-}5.55$\pm$1.10 & \hphantom{-}\textbf{5.58}$\pm$0.91 & \hphantom{-}5.56$\pm$0.84 \\
    yacht & 308 & 6 & -0.96$\pm$0.37 & -2.40$\pm$0.11 & \textbf{-0.59}$\pm$0.16 & -0.89$\pm$0.13 & -0.94$\pm$0.15 \\
    \bottomrule
  \end{tabular}
\bigskip
  \caption{Test data RMSE on UCI regression datasets. Bold numbers represent the best RMSEs.}
  \label{table:uci_small_rmse}
  \centering
  \setlength\tabcolsep{5pt}
  \begin{tabular}{lHHccccc}
    \toprule
    dataset & N & D & SGD & VI & SGD-PCA & AS & LIS \\
    \midrule
    boston & 506 & 13 & 3.50$\pm$0.98 & \textbf{3.47}$\pm$1.00 & \textbf{3.47}$\pm$0.96 & 3.54$\pm$0.97 & 3.58$\pm$0.94 \\
    concrete & 1030 & 8 & \textbf{5.18}$\pm$0.43 & 6.35$\pm$0.46 & 5.22$\pm$0.42 & 5.30$\pm$0.47 & 5.28$\pm$0.38 \\
    energy & 768 & 8 & \textbf{2.37}$\pm$0.24 & 2.66$\pm$0.25 & 2.39$\pm$0.24 & 2.45$\pm$0.23 & 2.48$\pm$0.24 \\
    naval & 11934 & 14 & .012$\pm$0.05 & .002$\pm$0.00 & .003$\pm$0.01 & \textbf{.001}$\pm$0.00 & .002$\pm$0.00 \\
    yacht & 308 & 6 & 0.89$\pm$0.29 & 1.99$\pm$0.55 & \textbf{0.68}$\pm$0.22 & 0.77$\pm$0.27 & 0.79$\pm$0.26 \\
    \bottomrule
  \end{tabular}
\end{table}
\begin{table}[ht!]
\fontsize{7.2}{9}\selectfont
  \caption{Test data calibration for UCI datasets. Bold numbers are closest to 95\% predictive coverage.}
  \label{table:uci_small_calibration}
  \centering
  \setlength\tabcolsep{5pt}
  \begin{tabular}{lHHccccc}
    \toprule
    dataset & N & D & SGD & VI & SGD-PCA & AS & LIS \\
    \midrule
    boston & 506 & 13 & 0.99$\pm$0.02 & 1.00$\pm$0.01 & 0.98$\pm$0.02 & 0.99$\pm$0.02 & \textbf{0.97}$\pm$0.02 \\
    concrete & 1030 & 8 & 0.85$\pm$0.03 & 0.98$\pm$0.01 & 0.89$\pm$0.03 & \textbf{0.92}$\pm$0.03 & 0.92$\pm$0.02 \\
    energy & 768 & 8 & 1.00$\pm$0.00 & \textbf{1.00}$\pm$0.01 & 1.00$\pm$0.00 & 1.00$\pm$0.00 & 1.00$\pm$0.00 \\
    naval & 11934 & 14 & \textbf{0.94}$\pm$0.12 & 0.99$\pm$0.01 & 0.99$\pm$0.04 & 0.99$\pm$0.01 & 0.99$\pm$0.01 \\
    yacht & 308 & 6 & \textbf{0.95}$\pm$0.07 & 1.00$\pm$0.00 & 0.98$\pm$0.03 & 0.99$\pm$0.02 & 0.99$\pm$0.02 \\
    \bottomrule
  \end{tabular}
\end{table}
The performance is evaluated by the test likelihoods (\autoref{table:uci_small_likelihoods}), RMSEs (\autoref{table:uci_small_rmse}), and calibration to 95\% credible intervals (\autoref{table:uci_small_calibration}). 

In test log-likelihood comparison, AS and LIS models outperform the SGD-PCA model while being comparable to deterministic and Bayesian NN. Deterministic NN provides a strong baseline for test RMSE and calibration results. SGD-PCA model performs better amongst subspace methods in test RMSEs. Our models are competitive with the SGD-PCA model in calibration results. This demonstrates the efficiency of our active subspace methods in capturing neural network output variability for effective Bayesian inference using AS or LIS of gradients obtained from random perturbations in the NN parameter space.

\vspace{-0.1in}
\section{Conclusion and Discussion} 
\vspace{-0.1in}
In this paper, we developed outcome-informed and likelihood-informed active subspace methods, where the constructed subspace effectively captures the NN output variability. This addresses the challenges arising from the high-dimensionality of typical neural network parameter space, thereby making standard Bayesian inference tractable. 
We demonstrated that our methods provide superior uncertainty estimates in the simulation study and comparable or better performance than the SGD-PCA baseline on log-likelihood, RMSE, and calibration metrics in UCI regression tasks. Our approach could be beneficial for UQ in hybrid simulations that combine physics with machine learning (e.g., \cite{Melland_2021}). Such simulations are often computationally expensive and can only be sampled over small ensemble sizes, which may warrant NN parameter space dimension reduction before conducting Bayesian inference.

\vspace{-0.1in}
\section*{Acknowledgments}
\vspace{-0.1in}

This research was supported by funding from the Advanced Scientific Computing Research program in the United States Department of Energy's Office of Science under projects $B\&R\#KJ0401010$ and $B\&R\#KJ0402010$.




\end{document}